\pdfoutput=1

\documentclass[11pt]{article}

\usepackage[preprint]{acl}

\usepackage{times}
\usepackage{latexsym}

\usepackage[T1]{fontenc}

\usepackage[utf8]{inputenc}

\usepackage{microtype}

\usepackage{inconsolata}

\usepackage{graphicx}

\usepackage{adjustbox,booktabs,multirow, amsmath}
\usepackage{xspace}
\usepackage{tcolorbox}
\newcommand{\method}{ChainEdit\xspace}
\newcommand{\eg}{\emph{e.g.}\xspace}

\newcommand{\vs}{\emph{vs.}\xspace}

\definecolor{mycolor}{RGB}{243,237,247}
\title{ChainEdit: Propagating Ripple Effects in LLM Knowledge Editing through Logical Rule-Guided Chains}

\author{Zilu Dong\thanks{Equal Contribution.},  Xiangqing Shen{\footnotemark[1]}, Zinong Yang, Rui Xia\thanks{Corresponding Author.}\\
        School of Computer Science and Engineering, \\ 
        Nanjing University of Science and Technology, China \\
        \{zldong, xiangqing.shen, znyang, rxia\}@njust.edu.cn}

\begin{document}

\maketitle

\begin{abstract}
Current knowledge editing methods for large language models (LLMs) struggle to maintain logical consistency when propagating ripple effects to associated facts. We propose \method, a framework that synergizes knowledge graph-derived logical rules with LLM logical reasoning capabilities to enable systematic chain updates. By automatically extracting logical patterns from structured knowledge bases and aligning them with LLMs' internal logics, \method dynamically generates and edits logically connected knowledge clusters. Experiments demonstrate an improvement of more than 30\% in logical generalization over baselines while preserving editing reliability and specificity. We further address evaluation biases in existing benchmarks through knowledge-aware protocols that disentangle external dependencies. This work establishes new state-of-the-art performance on ripple effect while ensuring internal logical consistency after knowledge editing. The code will be available at \url{https://github.com/NUSTM/ChainEdit}.
\end{abstract}

\section{Introduction}

Recent years have witnessed a continuous expansion of large language models' (LLMs) capabilities along with increasing model parameters. 
As new information constantly emerges and existing knowledge evolves, it is crucial to keep LLMs up-to-date and accurate. However, retraining these models to reflect such changes is prohibitively expensive and time-consuming due to their massive parameter sizes.
This highlights the importance of knowledge editing techniques for LLMs, which allow for targeted modifications without the need for full retraining, thus offering an efficient alternative to traditional methods.

\begin{figure}[t]
    \centering
    \includegraphics[width=\columnwidth]{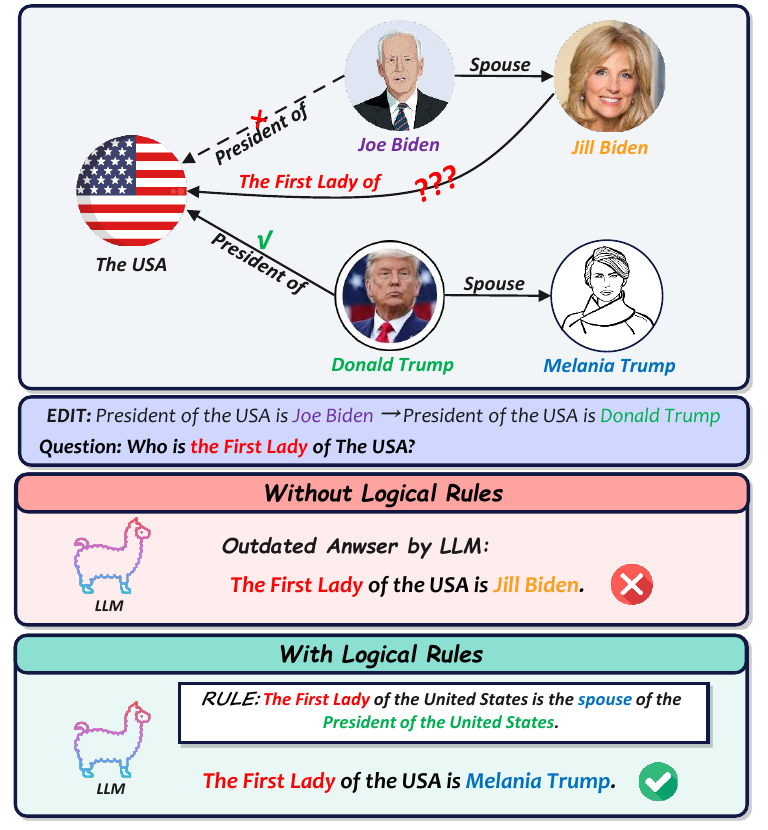}
                   \caption{After editing the president of USA, LLMs still use the original knowledge to answer the question about the first lady, instead of dynamically updating it using logical rules.}
    \label{fig:motivation}
\end{figure}

Knowledge editing approaches can be categorized into parameter‐preserving and parameter‐modifying paradigms \cite{DBLP:conf/emnlp/YaoWT0LDC023}. Parameter‐preserving methods, such as in‐context learning and memory‐based editing, guide model outputs via external inputs or auxiliary modules without touching the core weights. Parameter‐modifying methods adjust part of model parameters through constrained optimization and tailored loss functions. Both paradigms offer distinct trade-offs between edit durability and containment of side effects.

Despite their advances, existing knowledge editing approaches still face significant limitations, particularly regarding the impact of edited knowledge on model behavior. 
\citet{DBLP:journals/tacl/CohenBYGG24} identified a dual-dimensional ripple effect characterized by the necessary synchronization of logically associated knowledge and the preservation of unrelated knowledge stability, and in this work we focus on the synchronization challenge.

Existing studies on the ripple effect overlook the pivotal role of logical reasoning mechanisms. In real-world scenarios, updating one piece of knowledge can necessitate changes to other facts that are logically related. As illustrated in Figure~\ref{fig:motivation}, when editing the fact ``The US President is Donald Trump'', logically consistent generalization should simultaneously update ``The US First Lady is Melania Trump''. However, 
current methods do not make LLMs aware of the logical connection between US President and US First Lady, so they may still answer with the outdated information ``The first lady of the USA is Jill Biden.'' This type of ripple effect, referred to as ``logical generalization'', is driven by the capacity to infer which knowledge must also change based on logical relationships. 
This critical issue results in substantial deficiencies in the logical generalization ability of current model editing techniques, with logical generalization accuracy around 20\%, based on the \textsc{RippleEdits} benchmark proposed by \citet{DBLP:journals/tacl/CohenBYGG24}.

In essence, the need for logical generalization stems from humans’ logical reasoning mechanism—such as updating ``The US President'' should also update ``The US First Lady.'' While we cannot train LLMs to replicate human reasoning directly, we can distill the inference process into explicit logical rules and apply them during editing. As illustrated in Figure~\ref{fig:motivation}, applying the rule 
``The first lady of the United States is the spouse of the president of the United States.''  naturally leads to the correct behavior in the example. Moreover, although it has been found that LLMs demonstrate a latent understanding of logical rules \citep{DBLP:conf/acl/0009C23}, they currently fail to leverage this capability for logical generalization during knowledge editing.

To address these challenges, in this work we propose \method, which synergizes the intrinsic logical reasoning capacity of LLMs with logical rule systems to empower knowledge editing methods to generalize edits via logics.
Our key insight stems from the observation that both LLM knowledge editing and Knowledge Graph (KG) updating face similar challenges in chain updates. KG systems address this issue by applying logical rules to infer knowledge related to the update. Therefore, to enable LLMs to actively utilize their intrinsic logical reasoning capabilities, we integrate the rules mined from KGs with the logical reasoning of LLMs to construct a rule set that aligns with the internal logic of LLMs. This allows us to generate related knowledge influenced by a single updated piece of information, thus achieving the goal of chain updates.

Practically, \method operates through four phases: 1) extraction of candidate logical rules from KGs, 2) alignment of these rules with LLMs' internal logics, 3) dynamically generating affected knowledge queries based on editing rules, and 4) batch edits on both original and derived knowledge. This mechanism ensures deterministic updates for explicitly edited knowledge and its logical derivatives.

Our experiments adopt the \textsc{RippleEdits} benchmark, which is specifically designed to evaluate the ripple effects generated during model editing. When integrating \method with mainstream approaches like MEMIT, we observe substantial improvements in logical generalization metrics, for instance, achieving a 40.1\% (from 18.6\% to 58.7\%) improvement over baseline MEMIT while maintaining comparable performance on specificity metric constraints.

Moreover, we identify a critical limitation in existing benchmarks for assessing editing methods' generalization capabilities: their reliance on external KGs for question formulation. As their questions were formed using KG-derived knowledge, they requires external knowledge beyond the edited content to correctly answer, thereby obscuring the assessment of logical generalization capabilities. 
To address this, we propose an adaptation to the existing dataset, which consists of three variants: 1) select questions answerable with only the model's internal knowledge, 2) reconstruct the question by substituting intermediate knowledge with model-internal knowledge, and 3) embed intermediate knowledge required for the answer directly into the question itself. 
Our refined benchmark enables more accurate measurement of editing methods' true generalization capacity.

\section{Related Work}

\subsection{Knowledge Edit Methods}
Mainstream knowledge editing approaches can be broadly categorized into parameter-preserving and parameter-modifying paradigms \cite{DBLP:conf/emnlp/YaoWT0LDC023}. 
Our work focuses primarily on parameter-modifying methods, which inherently face greater challenges in controlling ripple effects compared to their parameter-preserving counterparts. 
These methods typically employ specialized loss functions and optimization constraints to control such ripple effects. For instance, MEND \cite{DBLP:conf/iclr/MitchellLBFM22} trained a meta-network with carefully designed contrastive samples that include both irrelevant knowledge and paraphrased generalizations, aiming to enhance generalization ability while preserving unrelated knowledge. Locate-and-edit approaches such as ROME \cite{DBLP:conf/nips/MengBAB22} and MEMIT \cite{DBLP:conf/iclr/MengSABB23} utilize causal mediation analysis to identify critical parameter layers, subsequently performing constrained optimization to update knowledge while maintaining original parameter space characteristics.

\subsection{Ripple Effects in Knowledge Edit}
\citet{DBLP:journals/tacl/CohenBYGG24} pioneered the formal characterization of ripple effects in model editing. Their benchmark framework introduced the first systematic evaluation protocol that simultaneously assesses both dimensions of ripple effects. Each knowledge edit undergoes rigorous evaluation through complementary metrics quantifying desired knowledge propagation and unintended side effects.

Recent advances address both aspects of ripple effects. \citet{DBLP:journals/corr/abs-2403-07825} focused on mitigating the harmful ripple effect of irrelevant knowledge. They proposed that knowledge with similar embeddings is more susceptible to the ripple effect and used this knowledge to assess the ripple effect generated in the latent space, which is difficult to detect. Conversely, \citet{DBLP:conf/emnlp/QinZHYLJ24} focused on analyzing beneficial ripple effects through GradSim, a gradient similarity measure between edited facts and associated knowledge. For in-context learning methods, \citet{DBLP:conf/emnlp/ZhaoYLC24} developed chain-of-thought strategies to amplify beneficial ripple effects.

Our work specifically targets ``Logical Generalization'', the most challenging category requiring models not only to memorize edited knowledge but also to logically integrate it with existing reasoning chains.

\begin{figure*}[t]
    \includegraphics[width=\linewidth]{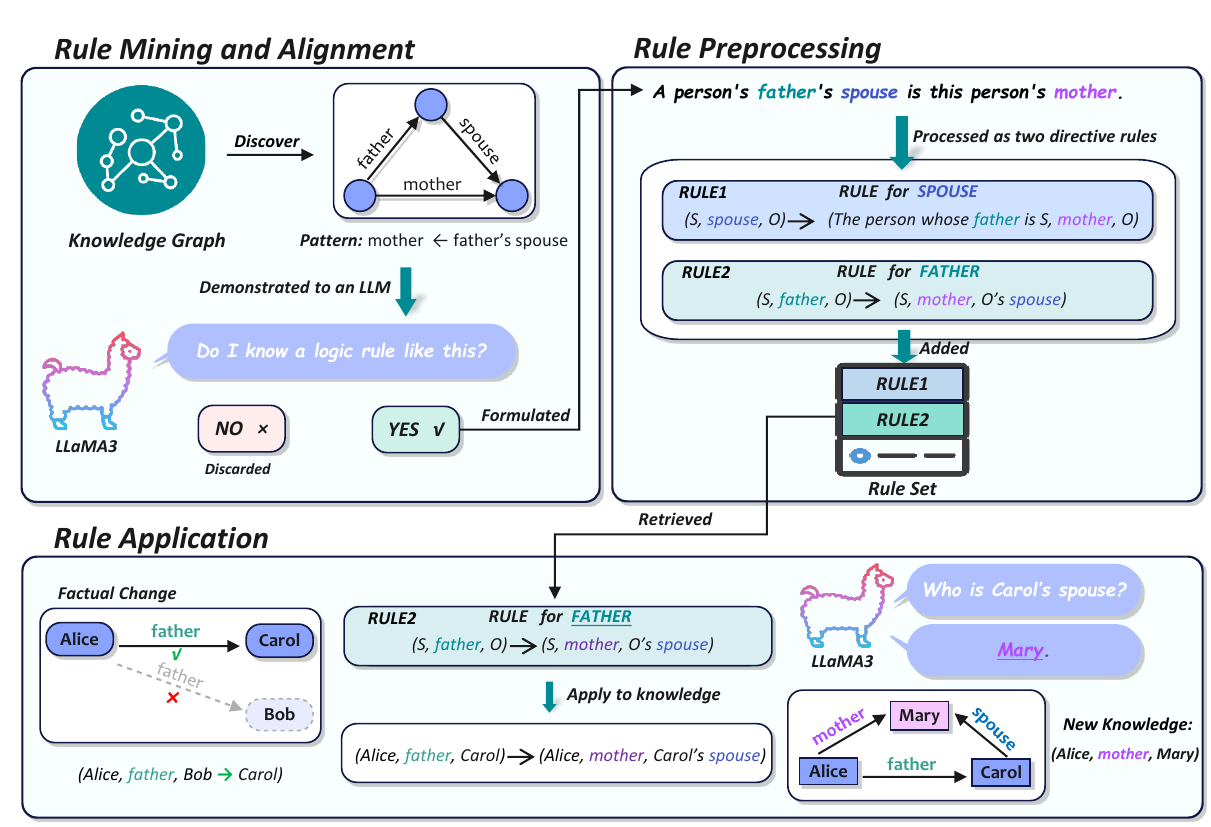}
    \caption{\method consists of three phases: (1) mine logical rule from KG and then align with LLM; (2) process the rule into directive rules (rule for edit) before adding into rule set; (3) retrieve and apply available rule to edit knowledge to generate relevant knowledge in the process of knowledge editing.}
    \label{fig:general_process}
\end{figure*}

\section{Methods}

Our proposed logic-aware knowledge editing framework (\method), inspired by KG update mechanisms, establishes a logical rule system to guide LLMs in synchronizing associated knowledge updates during editing. As depicted in Figure~\ref{fig:general_process}, our methods contains three phases: rule mining, preprocessing, and application.

\subsection{Rule Mining and Alignment}
To explicitly leverage LLMs' logical capabilities, we construct structured rules for direct application during knowledge editing. Our rule acquisition process comprises two stages:

\subsubsection{Rule Mining from KG}
We mine candidate logical rules from KG by identifying high-frequency alternative paths between relations. 

For a target relation $R$, we identify multihop relational paths $\langle r_1, ..., r_n \rangle$ that connect subject-object pairs within existing $R$-relation triples $(h,R,t)$. Formally, given a knowledge triple $(h,R,t) \in \mathcal{K}$, we discover alternative connection paths from knowledge graph  $\mathcal{K}$:
$h \xrightarrow{r_1} e_1 \xrightarrow{r2} \cdots \xrightarrow{r_3} t$
where the path $r_1 \circ \cdots \circ r_n$ provides logical support for the target relation $R$. Paths demonstrating statistically significant frequency (determined by co-occurrence count with the target relation) are retained as candidate rules.

For instance, analyzing ``Nationality''-relation triples reveals a high-frequency 2-hop path:
$h \xrightarrow{\text{BornIn}} m \xrightarrow{\text{CityOf}} t \Rightarrow h \xrightarrow{\text{Nationality}} t$.
This generates the rule $\texttt{Nationality} \leftarrow \texttt{(BornIn, CityOf)}$, indicating nationality derivation through birthplace.

\subsubsection{LLM-Rule Alignment}

Fundamentally, we aim for LLMs to leverage their internal logical rules during the editing process to ensure logical consistency in the edited knowledge. Therefore, it is essential to align the logic recognized by LLMs with the derived set of rules. Specifically, we transform candidate rule entries $R \leftarrow (r_1,...,r_n)$ into natural language statements and employ prompt design to enable LLMs to fully utilize their inherent logical reasoning capabilities to assess the universality of the rules (see Appendix~\ref{appendix:align_rules}). Rules with high universality are then incorporated into the rule set.

This alignment bridges symbolic rules with LLMs' logical reasoning, creating a rule base that respects LLMs' intrinsic logical structures while maintaining KG-derived constraints.

\subsection{Rule Preprocessing for Knowledge Editing}
We convert each extracted rule into one that directly guides the generalization of editing, named ``directive rule''. It is formalized as $\mathcal{R}: \langle \phi, \psi \rangle$, where $\phi$ denotes the trigger condition (logical predicate) and $\psi=(s_{new},r_{new},o_{new})$ specifies the knowledge generation template. Detailed syntax specifications appear in the Appendix~\ref{appendix:generation_structure}. 

Our rule templates explicitly handle multiple legitimate chain update paths under the same logical rule when knowledge modifications occur. Consider the rule $\texttt{father} \leftarrow \texttt{(sibling, father)}$, which indicates that the father of the sibling of a person should be their own father. When editing A's sibling relationship to B, this rule permits two distinct but logically valid update paths: updating A's father to match B's current father and adjusting B's father to align with A's existing father.
These two paths correspond to different real-world scenarios (\eg, A adopted by B's father \vs B adopted by A's father) that cannot be simultaneously true. Conventional rule representations struggle to capture such contextual variations, while our template system can explicitly encode both paths
\begin{equation}
\begin{aligned}
&\psi_1: (\text{Alice}, \text{father}, \text{Carol}.\text{father}) \\
&\psi_2: (\text{Carol}, \text{father}, \text{Alice}.\text{father})
\end{aligned}
\end{equation}.

\citet{DBLP:conf/emnlp/LiuYZLZSSJ24} have also discussed this logical path ambiguity issue and highlighted that, under conventional task setups, such ambiguity is unresolvable. Therefore, our rule template framework offers a more flexible and expressive mechanism that explicitly encodes both paths, enabling downstream systems or users to select the most contextually appropriate one.

\subsection{Rule Application for Relevant Knowledge Generation}
Given target edit $(s,r,o)$, our method first retrieves all matching rules $\mathcal{R}_i$ whose $\phi_i=r$, then performs variable substitution for each:
$\psi_i(s,o) = (s_{new}, r_{new}, o_{new})$ in rule $\mathcal{R}_i$. In other words, here we substitute the entity placeholder in $s_{new}$ and $o_{new}$ with real entities from the edit triple $(s,r,o)$.
For example, consider editing $(Alice,father,Carol)$ with triggered rule $\mathcal{R}: (S,father,O) \rightarrow (S,mother,O.spouse)$. Through variable substitution, we obtain the query item $q=(Alice,mother,Carol.spouse)$, then query LLM to get the entity representing $Carol.spouse$.

Structured prompting automates the knowledge derivation through the following operations: First, natural language queries are generated from formal query items $q_j=(s_j,r_j,o_j)$ - for instance, converting $Carol.spouse$ into the textual prompt \texttt{``The spouse of Carol is''}. This prompt is then fed to the LLM to elicit a response $v_j$, and to finally formulate a new knowledge triple $(s_j,r_j,v_j)$, as demonstrated by the example where the response \texttt{``Mary''} generates the associated triple $(Alice,mother,Mary)$.

Eventually, we perform batch editing on both original and derived knowledge using existing knowledge editing methods. This process ensures logical consistency in edits. For a complete example, see the Appendix~\ref{appendix:case_study}. Our framework demonstrates broad compatibility with existing editing methods, as evidenced by experiments across mainstream approaches in the next section.

\section{Experiments}

\subsection{Datasets and Metrics}
\label{sec:dataset}

\subsubsection{Existing Datasets}

To systematically evaluate the logical generalization capability of knowledge editing, we extend the ``\textsc{Popular}'' dataset from the \textsc{RippleEdits} benchmark proposed by \citet{DBLP:journals/tacl/CohenBYGG24} with enhanced evaluation protocols. The original dataset measures ripple effects through six metrics: Logical Generalization (LG), Compositionality I (CI), Compositionality II (CII), Subject Aliasing (SA), Forgetfulness (FF)\footnote{The ``Preservation'' metric in the original paper was renamed to Forgetfulness in their released dataset.}, and Relation Specificity (RS). Among these, LG, CI, CII, and SA assess the model's ability to propagate edits to logically related knowledge, while FF and RS assess unintended impacts on unrelated knowledge. Following \citet{DBLP:journals/corr/abs-2401-01286}'s KnowEdit benchmark, we combine CI and CII into a Reasoning (RE) metric, establishing a five-dimensional evaluation framework (LG/RE/SA/FF/RS). Furthermore, we use the ``reliability'' \cite{wang-etal-2024-easyedit} metric to assess the success rate of knowledge edits.

It should be noted that, although datasets such as MQUAKE \cite{DBLP:conf/emnlp/ZhongWMPC23} support the evaluation of multi-hop questions, they primarily focus on the concatenation reasoning of simple fact chains. In contrast, this study emphasizes the ability of LLMs to organically integrate edited knowledge with internal knowledge based on basic logic. Therefore, we did not conduct evaluations on such benchmarks.

\subsubsection{Our Dataset Variants}
While applying the \textsc{RippleEdits} benchmark, we identify a flaw in conventional evaluation protocols: their reliance on intermediate knowledge from external knowledge graphs that may misalign with LLMs' internal knowledge states. For instance, when evaluating the inference chain ``X graduated from Y University $\Rightarrow$ Y is located in Z City $\Rightarrow$ X's location is Z,'' discrepancies arise if the LLM internally believes ``Y University is in City E.'' Such misalignments conflate knowledge consistency with true logical generalization, as successful edits may still be penalized due to mismatched intermediate facts. This mismatch of intermediate knowledge leads the original dataset to measure, to some extent, the consistency between LLMs and external knowledge bases, rather than the true logical generalization ability of the editing method.

To address these limitations, we develop three LLM-adaptive dataset variants through methodological innovations. 
The \textbf{Filtered} Dataset implements a prescreening mechanism that removes questions requiring intermediate knowledge unverifiable by the LLM itself, preserving only queries resolvable through the model's internal knowledge post-editing. 
The \textbf{Replaced} Dataset dynamically regenerates intermediate knowledge using the LLM's current state, ensuring evaluation alignment with the model's evolving knowledge system. 
The \textbf{In-Prompt} Dataset employs context augmentation to explicitly inject intermediate knowledge into prompts, isolating logical reasoning capability assessment from knowledge completeness requirements. 

These variants serve distinct evaluation purposes: the first two diagnose the model's ability to generalize using its internal knowledge and edited knowledge, while the third evaluates enhanced reasoning when provided with auxiliary context. 
Through this taxonomy, our framework enables precise diagnosis of knowledge editing systems' strengths and limitations across different generalization scenarios.

\begin{table*}[th]
\centering
\small
\begin{tabular}{@{}lllllllll@{}}
\toprule
Model & Method                       & With Ours & Reliability & LG*  & RE   & SA   & RS   & FF   \\ \midrule
\multirow{8}{*}{\begin{tabular}[c]{@{}l@{}}Llama-3-\\      8B-Instruct\end{tabular}}  & \multirow{2}{*}{MEMIT} & w/ ours & 90.0 & 58.7 & 37.4 & 65.8 & 41.9 & 37.0 \\
      &                              & w/o ours  & 99.8        & 18.6 & 34.3 & 75.7 & 38.0 & 31.2 \\ \cmidrule(l){2-9} 
      & \multirow{2}{*}{MEMIT-Merge} & w/ ours   & 98.2        & 61.5 & 39.5 & 72.5 & 39.6 & 33.0 \\
      &                              & w/o ours  & 98.8        & 19.3 & 33.2 & 73.2 & 40.1 & 35.3 \\ \cmidrule(l){2-9} 
      & \multirow{2}{*}{FT}          & w/ ours   & 100.0       & 65.5 & 47.2 & 97.4 & 60.2 & 36.5 \\
      &                              & w/o ours  & 98.9        & 19.2 & 33.4 & 73.3 & 39.8 & 35.1 \\ \cmidrule(l){2-9} 
      & \multirow{2}{*}{LoRA}        & w/ ours   & 99.9        & 65.7 & 51.0 & 97.8 & 48.2 & 33.0 \\
      &                              & w/o ours  & 100.0       & 23.7 & 41.7 & 99.0 & 45.6 & 28.1 \\ \midrule
\multirow{8}{*}{\begin{tabular}[c]{@{}l@{}}Qwen2.5-\\      1.5B-Instruct\end{tabular}} & \multirow{2}{*}{MEMIT} & w/ ours & 87.4 & 55.8 & 26.2 & 53.3 & 41.1 & 35.1 \\
      &                              & w/o ours  & 99.0        & 22.1 & 24.0 & 59.7 & 40.6 & 30.0 \\ \cmidrule(l){2-9} 
      & \multirow{2}{*}{MEMIT-Merge} & w/ ours   & 96.6        & 61.5 & 29.5 & 59.0 & 35.8 & 29.4 \\
      &                              & w/o ours  & 99.0        & 22.3 & 24.1 & 60.6 & 38.6 & 29.9 \\ \cmidrule(l){2-9} 
      & \multirow{2}{*}{FT}          & w/ ours   & 100.0       & 55.6 & 43.9 & 97.9 & 36.3 & 27.1 \\
      &                              & w/o ours  & 100.0       & 16.5 & 43.5 & 98.6 & 42.6 & 31.7 \\ \cmidrule(l){2-9} 
      & \multirow{2}{*}{LoRA}        & w/ ours   & 100.0       & 59.6 & 49.0 & 99.1 & 33.6 & 24.1 \\
      &                              & w/o ours  & 100.0       & 17.6 & 39.5 & 99.1 & 29.1 & 19.4 \\ \bottomrule
\end{tabular}
\caption{The results (\%) obtained using the Qwen and Llama models on \textsc{RippleEdits}, comparing performance across different editing methods with and without our method. The metrics marked with * are the primary targets for improvement in this paper.  }
\label{tab:main-table}
\end{table*}

\subsection{Experimental Setup}
We conduct experiments using Llama-3-8B-Instruct \cite{llama3modelcard} and Qwen2.5-1.5B-Instruct \cite{qwen2.5,qwen2} models, implementing three editing methods: MEMIT, LoRA \cite{DBLP:conf/iclr/HuSWALWWC22}, MEMIT-Merge \cite{dong_memit-merge_2025}, and FT-M \cite{DBLP:journals/corr/abs-2401-01286}. All these evaluations adopt a single-instance editing protocol: after each knowledge edit, we immediately assess performance metrics defined in Section~\ref{sec:dataset} before reverting the edit to ensure isolation between test cases.

\subsubsection{Rule Mining Implementation}
Our systematic rule mining framework operates on Wikidata with three classes of rule targeting relations from the \textsc{RippleEdits} (\eg, nationality, kinship):

\begin{itemize}
\item \textbf{Inverse Relation}: For each target relation $r$, we sample 10,000 instances $(a \xrightarrow{r} b)$ and identify high-frequency inverse relations $r'$ from $b$ to $a$. Rules $r \leftrightarrow r'$ are established when the occurrence frequency exceeds threshold $\gamma$.

\item \textbf{Multi-Hop Path Mining}: For each $(a \xrightarrow{r} b)$ instance, we explore:
\begin{itemize}
\item 2-hop paths: $a \xrightarrow{r_1} m \xrightarrow{r_2} b$
\item 3-hop paths: $a \xrightarrow{r_1} m_1 \xrightarrow{r_2} m_2 \xrightarrow{r_3} b$
\end{itemize}
Paths with support count $\geq\gamma$ are retained as candidate rules $r \leftarrow (r_1,...,r_k)$. 
\end{itemize}
After deduplication, this process yields 3,120 candidate rules.

\begin{table*}[th]
\centering
\small
\begin{tabular}{@{}lllllllll@{}}
\toprule
Model & Method                          & Rule Set             & Reliability & LG*  & RE   & SA   & RS   & FF   \\ \midrule
\multirow{12}{*}{\begin{tabular}[c]{@{}l@{}}Qwen2.5-\\      1.5B-Instruct\end{tabular}} & \multirow{3}{*}{MEMIT} & pure mining & 88.7 & 54.7 & 25.9 & 54.1 & 39.5 & 33.7 \\
      &                              & with LLM           & 88.5        & 56.1 & 25.1 & 53.7 & 41.4 & 33.3 \\
      &                              & with LLM and human & 87.4        & 55.8 & 26.2 & 53.3 & 41.1 & 35.1 \\ \cmidrule(l){2-9} 
      & \multirow{3}{*}{MEMIT-Merge} & pure mining        & 97.1        & 53.0 & 29.8 & 63.5 & 36.3 & 30.5 \\
      &                              & with LLM           & 97.3        & 60.0 & 26.0 & 58.8 & 36.0 & 29.3 \\
      &                              & with LLM and human & 96.6        & 61.5 & 29.5 & 59.0 & 35.8 & 29.4 \\ \cmidrule(l){2-9} 
      & \multirow{3}{*}{FT}          & pure mining        & 100.0       & 50.5 & 42.3 & 98.1 & 37.0 & 29.0 \\
      &                              & with LLM           & 100.0       & 52.1 & 41.2 & 97.8 & 37.6 & 29.3 \\
      &                              & with LLM and human & 100.0       & 55.6 & 43.9 & 97.9 & 36.3 & 27.1 \\ \cmidrule(l){2-9} 
      & \multirow{3}{*}{LoRA}        & pure mining        & 100.0       & 63.0 & 45.8 & 98.4 & 33.9 & 23.9 \\
      &                              & with LLM           & 100.0       & 63.3 & 45.2 & 99.2 & 34.3 & 24.0 \\
      &                              & with LLM and human & 100.0       & 59.6 & 49.0 & 99.1 & 33.6 & 24.1 \\ \bottomrule
\end{tabular}
\caption{The result of applying \method using different rule sets: a) purely based on rule mining; b) method (a) plus LLM alignment; c) method (b) plus human selection.}
\label{tab:puremine-table}
\end{table*}

\subsection{Main Results}

As demonstrated in Table~\ref{tab:main-table}, the \method framework substantially enhances the logical generalization capabilities of mainstream batch editing methods.
In Llama-3-8B, MEMIT achieves a dramatic improvement in logical generalization from 18.6\% to 58.7\% , confirming that our chained rule trigger mechanism effectively establishes logical associations between knowledge items. Moreover, Qwen2.5-1.5B also exhibits significant LG improvements (22.1\% to 55.8\%), demonstrating \method's strong cross-architectural generalization. Table~\ref{tab:main-table} presents the editing results for single-instance scenarios. Additionally, we've included experiments on batch-instance editing in Appendix~\ref{appendix:batch_LRA_results}, which further confirms our method's broader applicability.

Although the edits of related knowledge increase editing complexity, the original knowledge editing success rate (reliability) of most methods remains stable. It should be noted that the original MEMIT method experienced a reliability drop of 9.8\% on the Llama model, which stems from the inherent flaw in its parameter update mechanism for editing multiple knowledge items with the same subject \cite{dong_memit-merge_2025}. By adopting the improved MEMIT-Merge strategy, the reliability was restored to 98.2\%, confirming that the performance fluctuation originated from the limitations of the method rather than the \method framework itself.

Specificity metrics show method-dependent fluctuations bounds within narrow bounds: MEMIT improves RS by 2.9\% while MEMIT-Merge decreases by 0.7\% with Llama. The minimal extent in decrease indicates that \method's precise boundary control prevents large-scale knowledge interference. Specifically, the RS metric indicates that, compared to Qwen2.5-1.5B, Llama-3-8B and other large models demonstrate a stronger ability to retain irrelevant knowledge, highlighting the positive impact of model capacity on stabilizing irrelevant knowledge.

Furthermore, contrary to initial expectations, the model scale shows limited correlation with the logical integration capability: Both large and small models exhibit similarly poor baseline LG performance (often <20\%) and comparable relative improvements after \method. Furthermore, even smaller models maintain stable RS metrics under our framework, proving \method's robustness across model sizes.

\begin{table*}[th]
\small
\centering
\begin{tabular}{@{}lllllllll@{}}
\toprule
Model & Dataset                    & With Ours & Reliability & LG*  & RE   & SA   & RS   & FF   \\ \midrule
\multirow{8}{*}{\begin{tabular}[c]{@{}l@{}}Llama-3-\\      8B-Instruct\end{tabular}}  & \multirow{2}{*}{original} & w/ ours & 98.2 & 61.5 & 39.5 & 72.5 & 39.6 & 33.0 \\
      &                            & w/o ours  & 98.8        & 19.3 & 33.2 & 73.2 & 40.1 & 35.3 \\ \cmidrule(l){2-9} 
      & \multirow{2}{*}{filtered}  & w/ ours   & 98.4        & 71.0 & 42.8 & 77.5 & 31.5 & 33.1 \\
      &                            & w/o ours  & 99.0        & 19.5 & 34.6 & 73.0 & 36.9 & 35.8 \\ \cmidrule(l){2-9} 
      & \multirow{2}{*}{replaced}  & w/ ours   & 96.8        & 65.8 & 38.5 & 70.4 & 37.8 & 32.2 \\
      &                            & w/o ours  & 99.1        & 19.6 & 30.7 & 70.3 & 39.6 & 34.5 \\ \cmidrule(l){2-9} 
      & \multirow{2}{*}{in-prompt} & w/ ours   & 97.7        & 67.6 & 69.9 & 81.8 & 39.4 & 34.1 \\
      &                            & w/o ours  & 99.0        & 33.1 & 68.4 & 83.8 & 40.9 & 35.7 \\ \midrule
\multirow{8}{*}{\begin{tabular}[c]{@{}l@{}}Qwen2.5-\\      1.5B-Instruct\end{tabular}} & \multirow{2}{*}{original} & w/ ours & 96.6 & 61.5 & 29.5 & 59.0 & 35.8 & 29.4 \\
      &                            & w/o ours  & 99.0        & 22.3 & 24.1 & 60.6 & 38.6 & 29.9 \\ \cmidrule(l){2-9} 
      & \multirow{2}{*}{filtered}  & w/ ours   & 97.3        & 72.5 & 30.8 & 59.3 & 27.8 & 29.5 \\
      &                            & w/o ours  & 99.0        & 17.5 & 21.7 & 60.5 & 29.0 & 30.9 \\ \cmidrule(l){2-9} 
      & \multirow{2}{*}{replaced}  & w/ ours   & 98.1        & 68.1 & 31.4 & 59.6 & 37.8 & 30.7 \\
      &                            & w/o ours  & 99.0        & 22.5 & 24.4 & 60.0 & 39.1 & 31.5 \\ \cmidrule(l){2-9} 
      & \multirow{2}{*}{in-prompt} & w/ ours   & 96.1        & 71.1 & 66.1 & 96.4 & 37.4 & 29.3 \\
      &                            & w/o ours  & 99.1        & 33.6 & 64.4 & 97.9 & 38.8 & 29.9 \\ \bottomrule
\end{tabular}
\caption{The results of applying \method in the MEMIT-Merge method on the original dataset and the variant datasets.}
\label{tab:various_dataset_results}
\end{table*}

\subsubsection{Ablation Study on Logical Rule Mining}
To validate the critical role of LLMs' internal logic, we compare two rule acquisition strategies: (1) pure KG-based mining \vs (2) KG mining with LLM's logical alignment. For fair comparison, we calibrate both rule sets to comparable sizes (around 180 rules). The LLM-filtered set retains 180 high-confidence rules through logical validation, while the pure KG set is subsampled via frequency thresholding to match this size.

As shown in the first two parts of Table~\ref{tab:puremine-table}, LLM-rule alignment outperforms pure KG mining, with logical generalization improving by 7\% (53\% → 60\%) on Llama-3-8B using MEMIT-Merge. This confirms that LLMs' internal reasoning patterns provide essential guidance for identifying practically applicable rules beyond statistical correlations.

Furthermore, we evaluate the impact of manual curation introduced in our full pipeline (primarily for computational efficiency). The last two sections of the table compare manually curated rules against fully automated mining, revealing negligible performance differences (LG: 60.0\% \vs 61.5\% with Llama-3-8B, MEMIT-Merge). This demonstrates that our core advantages stem from automated rule discovery rather than human intervention, establishing \method's scalability to large-scale applications.

\subsection{Datasets Integrating Edited and Internal Knowledge}

As detailed in Section~\ref{sec:dataset}, we construct three knowledge-adaptive datasets to eliminate evaluation biases caused by external knowledge dependencies. Table~\ref{tab:various_dataset_results} demonstrates MEMIT-Merge's performance across these datasets, with other edit methods' results provided in Appendix~\ref{appendix:detailed_datasets}.

\begin{figure}[th]
    \centering
    \includegraphics[width=\linewidth]{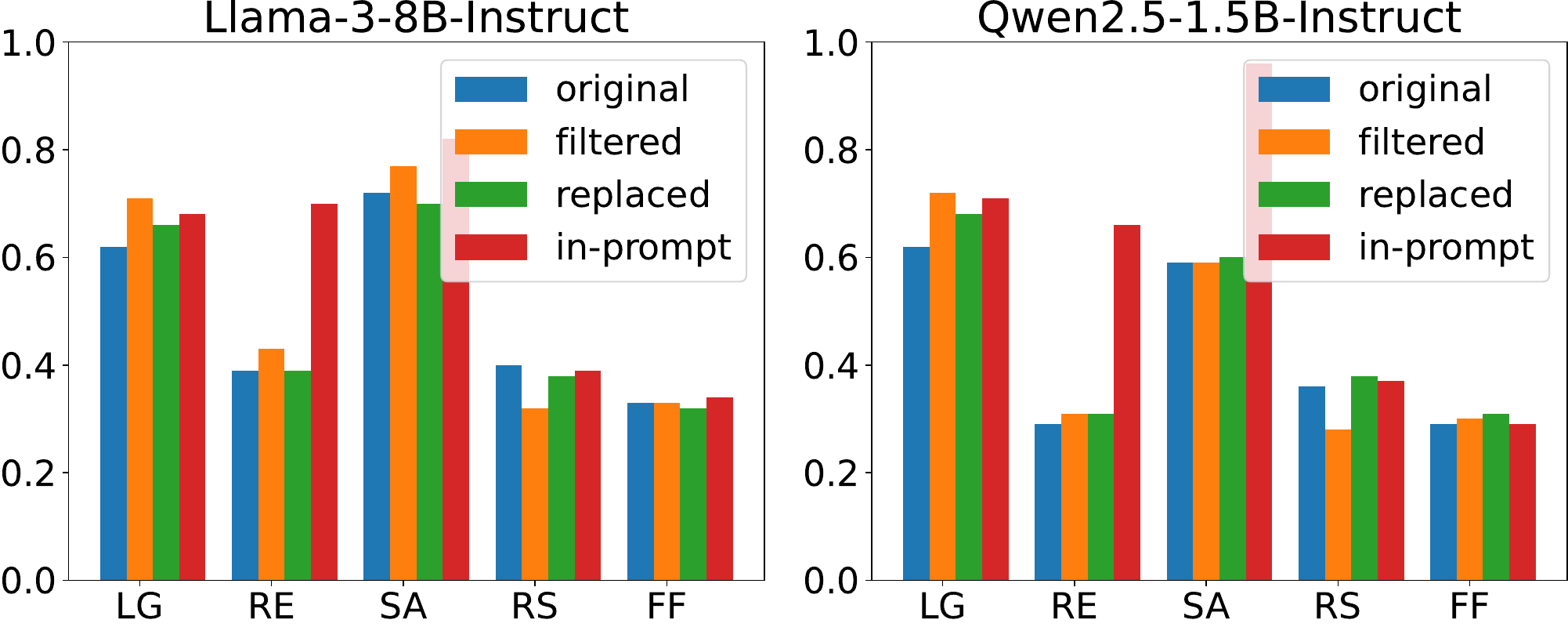}
    \caption{Comparison of MEMIT-Merge method results on the original dataset and our variant dataset after using \method.}
    \label{fig:plot_dataset}
\end{figure}

As shown in Table~\ref{tab:various_dataset_results}, after intermediate knowledge filtering, MEMIT-Merge method exhibits a more significant performance gap on the LG metric: without using \method, the LG value was only 19.5\%, while it surged to 71.0\% after applying \method in the Llama model. This phenomenon reveals that even if LLMs have already acquired intermediate knowledge, their knowledge system still cannot spontaneously establish logical associations with newly edited knowledge. As shown in Figure~\ref{fig:plot_dataset}, after using \method, the model's LG performance on the filtered dataset is generally better than that on the original dataset, indirectly confirming that in some cases, even if the model has successfully generalized the results, it cannot be accurately evaluated in the original data set derived from KG.

The replaced dataset dynamically substitutes KG-sourced intermediate knowledge with LLM-generated content. While this approach maintains question quantity (\vs reduction in filtered data), it introduces controlled noise through potential answer inaccuracies. As evidenced in Table~\ref{tab:various_dataset_results}, \method maintains robust LG performance (65.8\%) under these conditions, proving its capacity to leverage rule systems for reliable knowledge integration regardless of external answer correctness. This dual validation across filtered and replaced datasets establishes \method's unique ability to bridge fragmented internal knowledge through structured logical frameworks, outperforming conventional editing methods that rely solely on parametric memorization.

\begin{figure}[th]
    \centering
    \includegraphics[width=\linewidth]{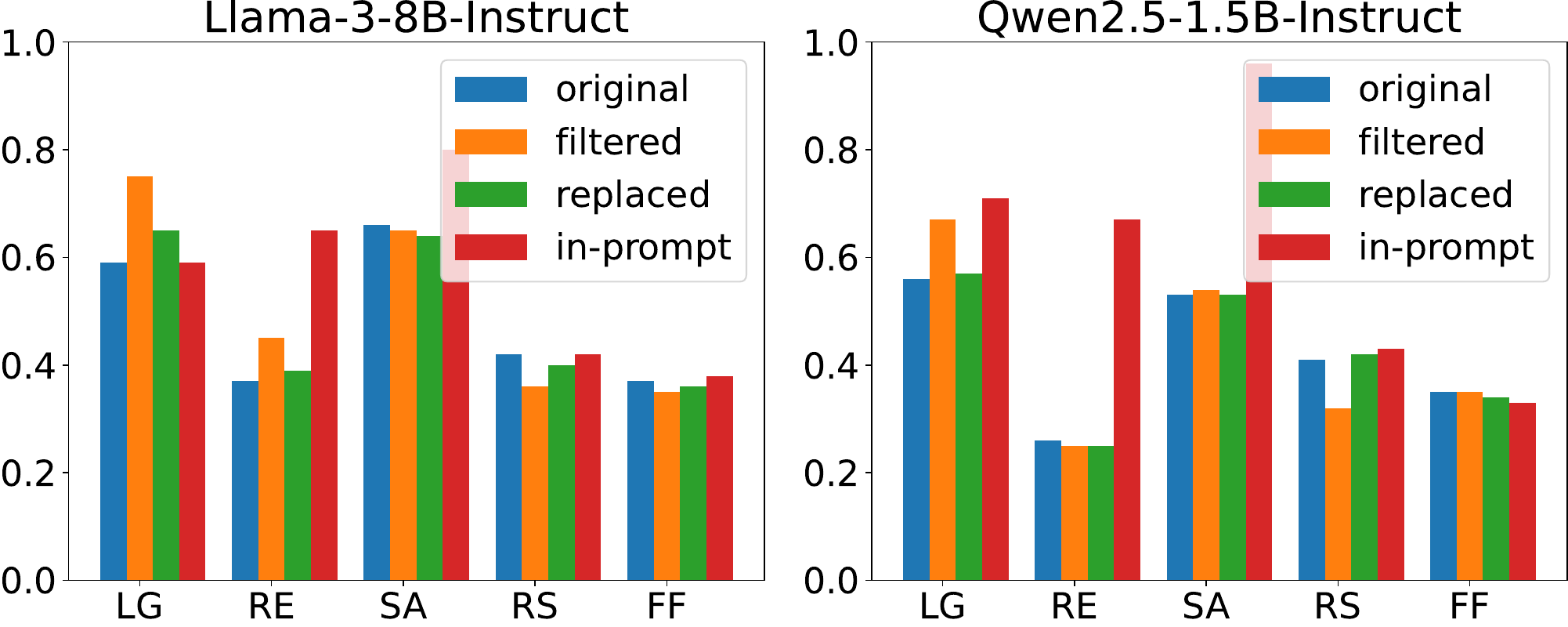}
    \caption{Comparison of MEMIT method results on the original dataset and our variant dataset after using \method.}
    \label{fig:plot_dataset2}
\end{figure}

\subsection{Integrating Edited Knowledge with Prompt-Provided Context}
This experiment explicitly injects intermediate knowledge into question contexts using the format: ``Given the following information: {intermediate knowledge}; Complete the following sentence: {original question}''. Unlike previous evaluations relying on internal knowledge integration, this setup tests LLMs' capacity to combine edited knowledge with external contextual information.

As shown in Table~\ref{tab:various_dataset_results}, \method enhances MEMIT-Merge's LG to 65.8\%, demonstrating effective multisource knowledge integration. Notably, SA scores exhibit anomalously high values due to simplified reasoning patterns: the injected context directly contains alias relationships, making correct answers trivially extractable.

Figure~\ref{fig:plot_dataset} and Figure~\ref{fig:plot_dataset2} reveals two critical insights: 1) LG and RE metrics significantly outperform original benchmark results when providing intermediate knowledge (LG: 33.1\% \vs 19.3\%, RE: 68.4\% \vs 33.2\% as shown in Table~\ref{tab:various_dataset_results}) without applying our method, confirming LLMs' stronger performance in combining edited knowledge with explicit context versus internal knowledge; 2) Despite these improvements, baseline LG scores remain suboptimal (<35\%), highlighting fundamental limitations in spontaneous logical rule application in LLM knowledge editing.

\section{Conclusion}
This paper addresses the ripple effect that occurs among interconnected facts during knowledge editing.
Starting from chain updates of KG, it expands the editing of a single piece of knowledge to the entire group of related knowledge using logical rules, thereby performing \method. On the \textsc{RippleEdits} benchmark, various editing methods 
integrating ChainEdit have shown a significant improvement in their ability in ``logical generalization''. Meanwhile, to address evaluation biases in existing benchmarks, we develop three LLM-adaptive dataset variants that decouple logical reasoning from external knowledge dependencies, where \method consistently shows superior generalization capability.

\section*{Limitations}
Our framework demonstrates promising results in improving the logical generalization ability of existing batch editing methods, yet several limitations warrant discussion to guide future research.

In the rule mining phase, reliance on knowledge graphs introduces limitations. The quality and comprehensiveness of rules may vary due to KGs' limited coverage and inability to capture all logical relationships.

Secondly, in the process of aligning the internal logic of large language models, the model's reasoning capability becomes a potential limiting factor. Particularly for models with smaller parameter scales and relatively weaker reasoning abilities, they may struggle to accurately understand or effectively apply certain complex rules, thereby impacting overall performance. 

Furthermore, in our rule mining strategy, we currently only retain rules that are applicable in most cases, which may overlook certain rules that are only valid under specific conditions. To address this challenge, conditional rule and selective application could be helpful.

Lastly, although our method demonstrates stability in editing specificity, as the rule set continues to expand, the number of edits may grow exponentially, which could adversely affect system performance. To address this challenge, future research could consider introducing optimization strategies such as rule priority mechanisms or rule application conditions.

\section*{Acknowledgments}

This work was supported by the Natural Science Foundation of China (No. 62476134).

\bibliography{references,  anthology, custom}
\newpage

\appendix
\section{Case study}
\label{appendix:case_study}

Consider the initial knowledge state  $\mathcal{K}_0$ containing the following triples:
\begin{equation}
\begin{aligned}
&\texttt{(Alice, father, Bob)} \\
&\texttt{(Alice, mother, Rose)} \\
&\texttt{(Bob, spouse, Rose)}
\end{aligned}
\end{equation}
When performing editing operation $\mathcal{E}$: \texttt{(Alice, father, Carol)}. Traditional editing methods only update the target triple, leading to logical inconsistencies. Specifically, after the edit, the model still treats \texttt{(Alice, mother, Rose)} as truth, whereas, according to the family relationship logic, Alice's mother should be updated to Carol's spouse.

By leveraging knowledge graph mining and LLM's knowledge, we obtain the following logical rules:
\begin{equation}
\mathcal{R}_1: \texttt{mother} \leftarrow (\texttt{father},\texttt{spouse})
\end{equation}, which represents that one's mother is the spouse of the father of this person.
This rule can be formulated into two directive rules: 
\begin{itemize}
\item  $\langle \phi: \texttt{father}, \psi: (S, mother, O.\texttt{spouse}) \rangle$
\item $\langle \phi: \texttt{spouse}, \psi: (X, mother, O) \rangle$, where X is the entity whose father is S
\end{itemize}

Edit operation $\mathcal{E}$ triggers following process:
\begin{enumerate}
\item \textbf{rule match}: detecting $\phi: \texttt{father}$ matching the directive rules of $\mathcal{R}_1$
\item \textbf{intermediate knowledge query}: perform query $q_1 = \texttt{spouse}(Carol,?)$ to LLM and thereby get response $v_1=Mary$
\item \textbf{relevant knowledge generation}: generate new knowledge triple $(Alice, mother,Mary)$
\item \textbf{batch edit}: form a edit batch ${\{\mathcal{E}, \texttt{(Alice, mother, Mary)}\}}$ and then perform batch editing.
\end{enumerate}

\section{Details of Aligning Logical Rules with the Internal Knowledge System of LLMs}
\label{appendix:align_rules}

To achieve alignment between logical rules and the internal knowledge system of LLMs, we designed a multi-stage semantic transformation process. First, we transform the formal rule $(A, r_1, B) \leftarrow (A, r_2, C)$ into a natural language proposition. The basic template is ``If the {relation} of {subject} is {object}, then...'', where nominal relations (such as nationality, occupation) adopt the structure ``the {relation} of {subject} is {object}'', while verbal relations (such as mentorship, employment) are adapted into the dynamic template ``{subject} {relation} {object}''. Taking the family relationship rule (A,father,B)←(A,mother,B.spouse) as an example, its natural language expression is ``If the father of A is B, then the mother of A is the spouse of B''.

The transformed natural language rules are input into the LLM for intermediate entity querying via a structured prompt. The model is required to provide its own logical rule to support or oppose the given rule and make a qualitative judgment on a five - level confidence scale (True/Usually True/Sometimes True/False/Uncertain).  (Example is given in ``Alignment Chating Example'' box) In the experiment, the first two levels are considered valid rules.

\begin{tcolorbox}[title = {Alignment Chating Example}]
[User] When the father of X is Y, then the sibling of X is the child of Y. 

[Assistant] \textit{The sibling of an person is his father's child.}   Answer: \textcolor{blue}{True} \\ 

[User] When the country of X is Y, then the continent of X is the continent of Y. 

[Assistant] \textit{The continent of a location is usually the same as the continent of the country location belongs to.} Answer: \textcolor{blue}{Usually True}
\end{tcolorbox}

\section{Structural Design of Knowledge Generation Templates in Rules}
\label{appendix:generation_structure}

This appendix provides a detailed explanation of the formal definition and execution mechanism of knowledge generation templates in logical rules. Given a rule $\mathcal{R}: \langle \phi, \psi \rangle$,where $\psi=(s_{new}, r_{new}, o_{new})$, the elements within the template are as follows:

$\phi$ (\textbf{Condition}): This represents the condition or premise of the rule. It specifies the circumstances under which the rule is applicable. For example, it could be a set of existing knowledge triples that must hold true for the rule to be triggered.

$\psi$ (\textbf{New knowledge}): This represents the new knowledge that is generated when the condition $\phi$ is satisfied. 

Triple entity placeholders, \eg :
\begin{itemize}
\item \texttt{S}: bind to current edit's subject entity
\item \texttt{O}: bind to current edit's object entity
\end{itemize}

Use the dot operator to represent multihop relation query, \eg :
\begin{itemize}
\item \texttt{O.father}: parsed as object entity's father
\item \texttt{S.birthplace.country}: nested operations is parsed from left to right, producing subject entity's place of birth's country of origin
\item \texttt{father.S}: parsed as the entity whose father is S 
\end{itemize}

\section{The results of ChainEdit in the batch edit scenario}
\label{appendix:batch_LRA_results}
In addition to single-instance editing scenarios, we conducted further experiments in batch edit scenarios. We tested our method's performance with both MEMIT-Merge and FT editing techniques across a batch size range from 2 to 100. The results are presented in \ref{tab:batch_LRA_results}.

When the batch size is increased, our method shows no significant change in reliability but a slight drop in LG. However, compared to the original method without our framework, it still performs at a very high level.

The decline tendency of the LG metric with increasing batch size is understandable. As the batch size increases, the number of edits grows, which inherently dilutes the effect of each individual edit to some extent. Since our method relies on the effectiveness of the editing process itself, the LG metric consequently decreases as the editing ability diminishes.

\begin{table*}[th]
\centering
\begin{tabular}{@{}lllllllll@{}}
\toprule
Method                       & Batchsize            & With Ours & Reliability & LG*  & RE   & SA   & RS   & FF   \\ \midrule
\multirow{8}{*}{MEMIT-Merge} & \multirow{2}{*}{2}   & w/ ours   & 97.6        & 61.0 & 29.3 & 60.0 & 37.1 & 30.6 \\
                             &                      & w/o ours  & 98.9        & 22.4 & 24.0 & 59.2 & 40.3 & 30.5 \\ \cmidrule(l){2-9} 
                             & \multirow{2}{*}{10}  & w/ ours   & 96.8        & 58.1 & 28.2 & 58.2 & 38.4 & 30.5 \\
                             &                      & w/o ours  & 98.8        & 20.6 & 24.5 & 60.4 & 39.9 & 31.2 \\ \cmidrule(l){2-9} 
                             & \multirow{2}{*}{50}  & w/ ours   & 94.8        & 56.1 & 29.2 & 58.0 & 35.3 & 30.5 \\
                             &                      & w/o ours  & 98.3        & 22.1 & 24.6 & 60.3 & 36.3 & 29.3 \\ \cmidrule(l){2-9} 
                             & \multirow{2}{*}{100} & w/ ours   & 91.5        & 52.1 & 28.6 & 54.6 & 35.3 & 29.6 \\
                             &                      & w/o ours  & 97.8        & 21.9 & 24.9 & 60.0 & 36.1 & 29.9 \\ \midrule
\multirow{8}{*}{FT}          & \multirow{2}{*}{2}   & w/ ours   & 99.8        & 58.3 & 41.9 & 95.4 & 26.7 & 20.7 \\
                             &                      & w/o ours  & 99.9        & 15.0 & 40.2 & 97.1 & 33.7 & 26.2 \\ \cmidrule(l){2-9} 
                             & \multirow{2}{*}{10}  & w/ ours   & 99.9        & 53.2 & 38.8 & 92.8 & 26.4 & 23.1 \\
                             &                      & w/o ours  & 100.0       & 14.7 & 35.3 & 92.6 & 26.6 & 21.2 \\ \cmidrule(l){2-9} 
                             & \multirow{2}{*}{50}  & w/ ours   & 99.3        & 52.4 & 35.3 & 86.6 & 18.6 & 17.8 \\
                             &                      & w/o ours  & 100.0       & 16.7 & 33.9 & 87.7 & 28.3 & 26.3 \\ \cmidrule(l){2-9} 
                             & \multirow{2}{*}{100} & w/ ours   & 97.6        & 50.7 & 31.0 & 82.5 & 11.3 & 11.6 \\
                             &                      & w/o ours  & 99.9        & 13.3 & 34.3 & 85.4 & 23.8 & 24.4 \\ \bottomrule
\end{tabular}
\caption{The results of ChainEdit in the batch edit scenario using Qwen2.5-1.5B-Instruct model on original dataset.}
\label{tab:batch_LRA_results}
\end{table*}

\section{Detailed Results in our datasets}
\label{appendix:detailed_datasets}

Tables~\ref{tab:llmfiltered_result}, \ref{tab:llmreplaced_result}, \ref{tab:midknow-table} present detailed results (\%) in the ``filtered'', ``replaced'', and ``in-prompt'' datasets, respectively.

\begin{table*}[th]
\centering
\begin{tabular}{@{}lllllllll@{}}
\toprule
Model & Method                       & With Ours & Reliability & LG*  & RE   & SA   & RS   & FF   \\ \midrule
\multirow{8}{*}{\begin{tabular}[c]{@{}l@{}}Llama-3-\\      8B-Instruct\end{tabular}}  & \multirow{2}{*}{MEMIT} & w/ ours & 88.0 & 74.8 & 45.3 & 65.3 & 35.9 & 34.5 \\
      &                              & w/o ours  & 99.1        & 19.3 & 33.0 & 72.4 & 35.0 & 35.2 \\ \cmidrule(l){2-9} 
      & \multirow{2}{*}{MEMIT-Merge} & w/ ours   & 98.4        & 71.0 & 42.8 & 77.5 & 31.5 & 33.1 \\
      &                              & w/o ours  & 99.0        & 19.5 & 34.6 & 73.0 & 36.9 & 35.8 \\ \cmidrule(l){2-9} 
      & \multirow{2}{*}{FT}          & w/ ours   & 100.0       & 78.4 & 44.2 & 98.9 & 47.5 & 34.8 \\
      &                              & w/o ours  & 100.0       & 13.9 & 40.6 & 97.5 & 69.3 & 41.0 \\ \cmidrule(l){2-9} 
      & \multirow{2}{*}{LoRA}        & w/ ours   & 99.9        & 75.7 & 51.3 & 98.5 & 45.4 & 32.1 \\
      &                              & w/o ours  & 100.0       & 17.2 & 37.1 & 98.8 & 45.2 & 28.6 \\ \midrule
\multirow{8}{*}{\begin{tabular}[c]{@{}l@{}}Qwen2.5-\\      1.5B-Instruct\end{tabular}} & \multirow{2}{*}{MEMIT} & w/ ours & 88.0 & 67.2 & 25.2 & 54.4 & 32.1 & 34.6 \\
      &                              & w/o ours  & 98.7        & 17.5 & 19.9 & 59.7 & 26.3 & 30.3 \\ \cmidrule(l){2-9} 
      & \multirow{2}{*}{MEMIT-Merge} & w/ ours   & 97.3        & 72.5 & 30.8 & 59.3 & 27.8 & 29.5 \\
      &                              & w/o ours  & 99.0        & 17.5 & 21.7 & 60.5 & 29.0 & 30.9 \\ \cmidrule(l){2-9} 
      & \multirow{2}{*}{FT}          & w/ ours   & 100.0       & 67.5 & 30.6 & 98.7 & 13.7 & 27.1 \\
      &                              & w/o ours  & 100.0       & 10.7 & 23.0 & 99.3 & 22.6 & 31.8 \\ \cmidrule(l){2-9} 
      & \multirow{2}{*}{LoRA}        & w/ ours   & 100.0       & 78.4 & 45.4 & 99.2 & 19.0 & 23.9 \\
      &                              & w/o ours  & 100.0       & 11.4 & 23.1 & 98.9 & 14.8 & 20.0 \\ \bottomrule
\end{tabular}
\caption{Detailed results on ``filtered'' dataset.}
\label{tab:llmfiltered_result}
\end{table*}

\begin{table*}[th]
\centering
\begin{tabular}{@{}lllllllll@{}}
\toprule
Model & Method                       & With Ours & Reliability & LG*  & RE   & SA   & RS   & FF   \\ \midrule
\multirow{8}{*}{\begin{tabular}[c]{@{}l@{}}Llama-3-\\      8B-Instruct\end{tabular}}  & \multirow{2}{*}{MEMIT} & w/ ours & 87.8 & 64.8 & 39.0 & 63.7 & 40.3 & 35.9 \\
      &                              & w/o ours  & 99.1        & 19.8 & 31.8 & 70.6 & 40.1 & 35.2 \\ \cmidrule(l){2-9} 
      & \multirow{2}{*}{MEMIT-Merge} & w/ ours   & 96.8        & 65.8 & 38.5 & 70.4 & 37.8 & 32.2 \\
      &                              & w/o ours  & 99.1        & 19.6 & 30.7 & 70.3 & 39.6 & 34.5 \\ \cmidrule(l){2-9} 
      & \multirow{2}{*}{FT}          & w/ ours   & 100.0       & 64.6 & 45.2 & 95.1 & 53.9 & 32.5 \\
      &                              & w/o ours  & 100.0       & 13.9 & 40.6 & 97.5 & 69.3 & 41.0 \\ \cmidrule(l){2-9} 
      & \multirow{2}{*}{LoRA}        & w/ ours   & 100.0       & 66.3 & 49.3 & 95.4 & 45.7 & 32.0 \\
      &                              & w/o ours  & 100.0       & 17.2 & 37.1 & 98.8 & 45.2 & 28.6 \\ \midrule
\multirow{8}{*}{\begin{tabular}[c]{@{}l@{}}Qwen2.5-\\      1.5B-Instruct\end{tabular}} & \multirow{2}{*}{MEMIT} & w/ ours & 87.8 & 56.8 & 25.3 & 53.5 & 41.5 & 33.8 \\
      &                              & w/o ours  & 99.1        & 24.0 & 24.2 & 58.5 & 38.2 & 30.7 \\ \cmidrule(l){2-9} 
      & \multirow{2}{*}{MEMIT-Merge} & w/ ours   & 98.1        & 68.1 & 31.4 & 59.6 & 37.8 & 30.7 \\
      &                              & w/o ours  & 99.0        & 22.5 & 24.4 & 60.0 & 39.1 & 31.5 \\ \cmidrule(l){2-9} 
      & \multirow{2}{*}{FT}          & w/ ours   & 100.0       & 56.4 & 42.8 & 95.6 & 35.4 & 27.7 \\
      &                              & w/o ours  & 100.0       & 19.4 & 40.7 & 96.5 & 42.3 & 31.8 \\ \cmidrule(l){2-9} 
      & \multirow{2}{*}{LoRA}        & w/ ours   & 100.0       & 59.8 & 49.0 & 95.2 & 34.2 & 24.1 \\
      &                              & w/o ours  & 100.0       & 19.8 & 35.9 & 94.1 & 31.5 & 17.1 \\ \bottomrule
\end{tabular}
\caption{Detailed results on ``replaced'' dataset.}
\label{tab:llmreplaced_result}
\end{table*}

\begin{table*}[th]
\centering
\begin{tabular}{@{}lllllllll@{}}
\toprule
Model & Method                       & With Ours & Reliability & LG*  & RE   & SA   & RS   & FF   \\ \midrule
\multirow{8}{*}{\begin{tabular}[c]{@{}l@{}}Llama-3-\\      8B-Instruct\end{tabular}}  & \multirow{2}{*}{MEMIT} & w/ ours & 86.3 & 59.5 & 64.6 & 80.2 & 42.2 & 38.1 \\
      &                              & w/o ours  & 99.0        & 32.1 & 68.1 & 84.9 & 39.7 & 35.2 \\ \cmidrule(l){2-9} 
      & \multirow{2}{*}{MEMIT-Merge} & w/ ours   & 97.7        & 67.6 & 69.9 & 81.8 & 39.4 & 34.1 \\
      &                              & w/o ours  & 99.0        & 33.1 & 68.4 & 83.8 & 40.9 & 35.7 \\ \cmidrule(l){2-9} 
      & \multirow{2}{*}{FT}          & w/ ours   & 100.0       & 71.7 & 65.4 & 89.7 & 58.9 & 36.1 \\
      &                              & w/o ours  & 100.0       & 27.0 & 66.8 & 88.6 & 70.3 & 40.4 \\ \cmidrule(l){2-9} 
      & \multirow{2}{*}{LoRA}        & w/ ours   & 99.7        & 69.3 & 68.4 & 94.5 & 49.1 & 34.0 \\
      &                              & w/o ours  & 100.0       & 26.8 & 62.6 & 95.9 & 46.2 & 29.2 \\ \midrule
\multirow{8}{*}{\begin{tabular}[c]{@{}l@{}}Qwen2.5-\\      1.5B-Instruct\end{tabular}} & \multirow{2}{*}{MEMIT} & w/ ours & 85.5 & 71.3 & 66.8 & 96.0 & 42.8 & 33.3 \\
      &                              & w/o ours  & 98.9        & 35.6 & 62.4 & 97.8 & 38.9 & 29.2 \\ \cmidrule(l){2-9} 
      & \multirow{2}{*}{MEMIT-Merge} & w/ ours   & 96.1        & 71.1 & 66.1 & 96.4 & 37.4 & 29.3 \\
      &                              & w/o ours  & 99.1        & 33.6 & 64.4 & 97.9 & 38.8 & 29.9 \\ \cmidrule(l){2-9} 
      & \multirow{2}{*}{FT}          & w/ ours   & 100.0       & 59.5 & 57.9 & 94.7 & 37.1 & 27.4 \\
      &                              & w/o ours  & 100.0       & 22.5 & 57.6 & 96.2 & 41.9 & 30.7 \\ \cmidrule(l){2-9} 
      & \multirow{2}{*}{LoRA}        & w/ ours   & 100.0       & 57.5 & 50.2 & 99.3 & 34.6 & 26.4 \\
      &                              & w/o ours  & 100.0       & 17.6 & 39.5 & 99.1 & 29.1 & 19.4 \\ \bottomrule
\end{tabular}
\caption{Detailed results on ``in-prompt'' dataset.}
\label{tab:midknow-table}
\end{table*}

\end{document}